\documentclass[twoside]{article}

\usepackage{hyperref}
\hypersetup{colorlinks = true, citecolor = blue, linkcolor=blue}
\usepackage[utf8]{inputenc} 
\usepackage[T1]{fontenc}    
\usepackage{url}            
\usepackage{booktabs}       
\usepackage{amsfonts}       
\usepackage{nicefrac}       
\usepackage{microtype}      
\usepackage{xcolor}         
\usepackage{amssymb}
\usepackage{amsmath}
\usepackage{amsthm}
\usepackage[linesnumbered,ruled]{algorithm2e}
\usepackage{algpseudocode}
\usepackage{booktabs}
\usepackage[shortlabels]{enumitem}
\usepackage{mathtools}
\usepackage{subfigure}
\usepackage{mdframed}
\usepackage{courier}
\usepackage[square,numbers, sort&compress]{natbib}
\newcommand{\mycomment}[1]{}

\usepackage[accepted]{aistats2023}
\begin{document}

%

%

\twocolumn[

\aistatstitle{Conformalized \textbf{\textit{Unconditional}} Quantile Regression}
\aistatsauthor{ Ahmed M. Alaa \And Zeshan Hussain \And  David Sontag}

\aistatsaddress{UC Berkeley \And  MIT \And MIT} ]

\begin{abstract}
We develop a predictive inference procedure that combines conformal prediction (CP) with {\it unconditional} quantile regression (QR)---a commonly used tool in econometrics \cite{firpo2009unconditional} that involves regressing the {\it recentered influence function}~(RIF)~of the quantile functional over input covariates. Unlike the more widely-known conditional QR, unconditional QR explicitly captures the impact of changes in covariate distribution on the quantiles of the marginal distribution of outcomes. Leveraging this property, our procedure issues adaptive predictive intervals with localized frequentist coverage guarantees. It operates by fitting a machine learning model for the RIFs using training data, and then applying the CP procedure for any test covariate with respect to a ``hypothetical'' covariate distribution {\it localized} around the new instance. Experiments show that our procedure is adaptive to heteroscedasticity, provides transparent coverage guarantees that are relevant to the test instance at hand, and performs competitively with existing methods in terms of efficiency. 
\end{abstract}

\section{INTRODUCTION}
\label{Sec1}
Consider a training data set $(X_1, Y_1), \ldots, (X_n, Y_n)$, and a test point $(X_{n+1}, Y_{n+1})$, with the training and test data all drawn independently from the same distribution, i.e., 
\begin{align}
(X_i, Y_i) &\overset{i.i.d}{\sim} P = P_X \times P_{Y|X},\, i=1,\ldots, n + 1. 
\label{eq1}
\end{align}
Here, each $X_i \in \mathbb{R}^d$ is a covariate vector, while $Y_i \in \mathbb{R}$ is a response variable. The joint distribution over covariates and responses, $P$, is unknown. In this paper, we tackle the problem of {\it predictive inference}, where given the covariate vector $X_{n+1}$ for a new test point, the goal is to construct a predictive interval that is likely to contain the true response $Y_{n+1}$ with probability at least $1-\alpha$, for some $\alpha \in (0,1)$. More precisely, our goal is to use the $n$ training sample points to construct a set-valued function:
\begin{align}
\widehat{C}_n(x) \coloneqq \widehat{C}_n((X_1, Y_1), \ldots, (X_n, Y_n), x) \subseteq \mathbb{R},
\label{eq2}
\end{align}
such that for a new test point $(X_{n+1}, Y_{n+1}) \sim P$, the response $Y_{n+1}$ falls in \mbox{\footnotesize $\widehat{C}_n(X_{n+1})$} with probability $1-\alpha$. Intervals satisfying this {\it coverage} condition are said to be {\it valid}. 

The sense in which predictive inferences are valid determines the relevance of the corresponding coverage guarantees to specific prediction instances. The weakest form of validity is when predictive intervals cover the true response on average---such intervals are said to be {\it marginally valid}. Formally, marginal validity is satisfied when
\begin{equation} 
\mathbb{P}\left[\,Y_{n+1} \in \widehat{C}_{n}(X_{n+1}) \,\right] \geq 1-\alpha,
\label{eq3r}
\end{equation}
where the probability is defined with respect to randomness of both training and testing data. The coverage condition in (\ref{eq3r}) is said to be {\it distribution-free} if it holds for all $P$. Conformal prediction (CP), described formally in Section~\ref{Sec2}, is a popular framework for predictive inference that guarantees distribution-free marginal validity in finite samples \cite{vovk1999machine, vovk2005algorithmic, papadopoulos2008inductive, lei2013distribution, lei2014distribution, lei2018distribution, romano2019conformalized}. In its most basic form, CP achieves marginal validity by issuing a fixed-length interval for all prediction instances.

In many applications, it is important to ensure {\it transparency} in communicating uncertainty in predictions issued for individual users. For instance, suppose that $X_i$ is a set of risk factors for patient~$i$ (e.g., age, blood pressure, etc.), and $Y_i$ is a measure of kidney function (e.g., eGFR). For a new patient, our goal would be to predict a range of values for their future eGFR with a predetermined degree of confidence, i.e., we would like to be able to make a statement along the lines of: ``Based on your risk factors, there is a 95$\%$ chance that your eGFR will decline by 1.8–5.2 mL/min over the next 3 years''. Marginal coverage guarantees that predicted ranges are accurate for 95$\%$ of the patients on average, but can be arbitrarily inaccurate for specific prediction instances. Since marginal coverage is defined with respect to $P_X$, we expect coverage to be violated in regions~with~few training examples in covariate-space, i.e., instances for which the predictive model has high {\it epistemic} uncertainty.~Hence,~the marginally-valid fixed-length intervals issued by vanilla CP may not be informative for prediction instances~to~which~uncertainty quantification matters the most.

{\bf Adaptive and transparent CP.} In this paper, we address the following question: how can we communicate model uncertainty in specific prediction instances in an {\it adaptive} and {\it transparent} manner? That is, we would like to construct a predictive inference procedure that {\it adapts} the length of its issued intervals based on the varying level of uncertainty across different prediction instances, and reports a coverage guarantee that is ``relevant'' to each specific instance. Ideally, we would like to develop a predictive inference procedure that achieves the following conditional coverage guarantee: 
\begin{equation}
\mathbb{P}\left[\,Y_{n+1} \in \widehat{C}_{n}(X_{n+1}) \,\big|\, X_{n+1}=x\,\right] \geq 1-\alpha,\,
\label{eq3}
\end{equation}
for almost all\footnote{We write ``almost all $x$'' to mean that the set of points where the bound fails has measure zero under $P_X$.} $x \in \mathbb{R}^d$. Predictive intervals that satisfy (\ref{eq3}) are said to be {\it conditionally valid}.~A~procedure~that~satisfies (\ref{eq3}) is transparent as its guarantee holds for each prediction instance, and is adaptive if the length of \mbox{\small $\widehat{C}_{n}$} for a given $x$ reflects the relative level of uncertainty in this specific instance. It is known that distribution-free validity in the sense of (\ref{eq3}) is impossible to achieve for non-trivial predictions \cite{vovk2012conditional, lei2014distribution, foygel2021limits}. Hence, we build on the CP framework to develop an adaptive predictive inference procedure that satisfies an approximate version of (\ref{eq3}), and transparently reports the granularity of coverage for each prediction instance. 

For each new test point, our procedure reports an instance-specific predictive interval and identifies a local region containing the instance $X_{n+1}=x$, over which the procedure is marginally valid, i.e., the inference is reported as follows:

\vspace{.1in}
\begin{mdframed} 
\vspace{.05in}
{\color{black}{For $X_{n+1}=x$, report $\widehat{C}_{n}(x), \widehat{\mathcal{S}}_n(x)$ such that:
\begin{equation}
\mathbb{P}\left[\,Y \in \widehat{C}_{n}(X) \,\big|\, X \in \widehat{\mathcal{S}}_n(x)\,\right] \geq 1-\alpha,\,\, \forall  x \in \mathbb{R}^d, \nonumber
\label{eq3proposed}
\end{equation}
where $\widehat{\mathcal{S}}_n(x) \subseteq \mathbb{R}^d$, which we call a {\it relevance subgroup}, is a local region containing $x$. 
\vspace{.05in}
}}
\end{mdframed}
\vspace{.025in}

In the clinical example discussed earlier, our procedure would communicate uncertainty with an individual patient as follows: ``The model predicts that your eGFR will decline by 1.8–5.2 mL/min over the next 3 years. The predictions of the model tend to be accurate 95$\%$ of the time for patients {\it similar} to you defined by the patient subgroup \mbox{\small $\widehat{\mathcal{S}}_n(x)$}, but the accuracy will vary from one patient to another within this group''. By communicating uncertainty in this more transparent form, the clinician can reason about the relevance~of this prediction to the patient at hand by inspecting the reported subgroup, e.g., checking if the relevance subgroup includes patients with different disease phenotypes.

The key idea behind our method is to view localized predictive inference as a marginal CP problem under a ``hypothetical'' covariate distribution $G_x$ localized~around~the~test point $x$ instead of the true distribution $P_X$. This is implemented through a commonly used approach in econometrics, known as unconditional quantile regression (UQR), which estimates the marginal quantiles of outcomes within arbitrary local subsets (i.e., relevance subgroups) of the covariate space \cite{firpo2009unconditional}. It does so by regressing the recentered influence function (RIF) of the quantile functional over covariates, and marginalizing the predicted RIFs within relevance subgroups. This is different from~conditional~quantile regression \cite{romano2019conformalized}, for which the regression targets~do~not~recover marginal quantiles when averaged over subgroups.

Our procedure involves two steps. First, we use the UQR model to generate a nested sequence of predictive~bands~for each relevance subgroup. Next, we select the tightest band that achieves coverage within each subgroup using a held-out calibration set. In the rest of the paper,~we~explain~the two steps of our procedure; we first start by~providing~a~brief background on the standard CP method in the next Section. 

\section{Conformal Prediction}
\label{Sec2}
The standard {\it split} CP procedure relies on sample splitting for constructing predictive intervals that satisfy finite-sample coverage guarantees \cite{vovk1999machine,vovk2005algorithmic, papadopoulos2008inductive}. Assuming that all data points are exchangeable, the procedure splits the data set into two disjoint subsets: a proper training set $\{(X_i, Y_i): i \in \mathcal{D}_t\}$, and a {\it calibration} set $\{(X_i, Y_i): i \in \mathcal{D}_c\}$. Then, a machine learning model $\widehat{\mu}(x)$ is fit to the training data set $\mathcal{D}_t$, and a {\it conformity score} $V (.)$ is computed for all samples in $\mathcal{D}_c$---this score measures how unusual the prediction looks relative to previous examples.~A~typical~choice~of~$V (.)$ is the absolute residual, i.e., $V(x, y) \coloneqq |\,\widehat{\mu}(x) - y\,|$. The conformity scores are evaluated as follows:
\begin{equation}
V_i \coloneqq V(X_i, Y_i) = |\,\widehat{\mu}(X_i) - Y_i\,|,\, \forall i \in \mathcal{D}_c.
\label{eq6}
\end{equation}
For a given miss-coverage level $\alpha$, we then compute a quantile of the empirical distribution of the absolute residuals,
\begin{equation}
Q_\mathcal{V}(1-\alpha) \coloneqq (1-\alpha)(1+1/|\mathcal{D}_c|) \mbox{-th quantile of $\mathcal{V}$,}  
\label{eq7}
\end{equation}
where $\mathcal{V} = \{V_i: i \in \mathcal{D}_c\}$. Finally, the prediction interval at a new point $X_{n+1}=x$ is given by
\begin{equation}
\widehat{C}_n(x) = [\,\widehat{\mu}(x) - Q_\mathcal{V}(1-\alpha),\,\widehat{\mu}(x) + Q_\mathcal{V}(1-\alpha)\,].
\label{eq8}
\end{equation}
The CP intervals have a fixed length of $2Q_\mathcal{V}(1-\alpha)$, independent of $X_{n+1}$, which is sufficient for satisfying marginal validity but does not adapt to the varying degrees of uncertainty across different prediction instances.

\section{Conformalized Unconditional Quantile Regression (CUQR)} 
In this Section, we describe the two steps involved in our procedure, which we call conformalized unconditional quantile regression (CUQR). Indeed, ours is not the first adaptive variant of CP---we compare our method with existing approaches to adaptive uncertainty quantification~in~Section~\ref{Sec4}.

\subsection{Step 1: Unconditional Quantile Regression (UQR)} 
\label{sec31}
Consider the true populational counterpart~of~\mbox{\footnotesize $\widehat{C}_n$} in (\ref{eq3}), i.e.,
\begin{equation}
C^*(x) \coloneqq \left[Q\left(\alpha/2, x\right), Q\left(1-\alpha/2, x\right)\right],
\label{eq4}
\end{equation}
where $C^*$ is the predictive interval given {\it oracle} knowledge of $P$, $Q(\alpha, x)$ is the level-$\alpha$ quantile of $Y|X=x$, i.e., $Q(\alpha, x) \coloneqq \inf\{y \in \mathbb{R}: F(y\,|\,X=x) \geq \alpha\}$, and $F(.)$ is the conditional cumulative density function (CDF), $F(y\,|\,X=x) \coloneqq \mathbb{P}(Y \leq y\,|\,X=x)$. By definition, the oracle band in (\ref{eq4}) is conditionally valid in the~sense~of~(\ref{eq3}). Hence, a sensible guess of an uncertainty~band~that is both adaptive and transparent can be obtained by directly estimating the conditional quantile $Q(., x)$.

{\bf Nested sequence of plug-in estimates.}

We use a {\it plug-in} approach for estimating $C^*$ by replacing the conditional quantile in (\ref{eq4}) with a consistent~estimate~$\widehat{Q}$, i.e., $\widehat{C}_n(x) = [\,\widehat{Q}(\alpha/2, x), \widehat{Q}\left(1-\alpha/2, x\right)\,]$. While plug-in models can learn accurate estimates of $C^*$, they do not provide finite-sample coverage guarantees. To take advantage of both the adaptivity of plug-in estimates and the finite-sample coverage of the CP framework, a typical approach is to ``conformalize'' these plug-in estimates \cite{romano2019conformalized}. In what follows, we explain how our procedure creates conformity scores based on plug-in estimates of $C^*$.

Instead of constructing predictive intervals using a point estimate of $Q(\alpha, .)$ obtained from a single plug-in model $\widehat{Q}(\alpha, .)$, we generate a set of ``candidate'' estimates of the conditional quantile function and use the calibration set $\mathcal{D}_c$ to pick the narrowest candidate band that achieves the desired coverage. More precisely, we define a set of predictive intervals for covariate $x$, $\widehat{\mathcal{C}}(x)$, as follows:
\begin{equation}
\widehat{\mathcal{C}}(x) \coloneqq \{\widehat{\mathcal{C}}_{\tilde{\alpha}}(x) = \widehat{\mu}(x) \pm \widehat{Q}(\tilde{\alpha}, x)\}_{\tilde{\alpha} \in (0, 1)},
\label{eq4nested}
\end{equation}
where $\widehat{Q}(\tilde{\alpha}, x)$ is a plug-in estimate of the level-$\tilde{\alpha}$ conditional quantile of the {\bf model residual}~at~$x$. We require that the plug-in estimates are monotonic: $\widehat{Q}(\tilde{\alpha}, x) \leq \widehat{Q}(\tilde{\alpha}^\prime, x)$ for $\tilde{\alpha} \leq \tilde{\alpha}^\prime$, i.e., no quantile crossing. Thus, $\widehat{\mathcal{C}}(x)$ comprises a nested sequence of candidate intervals, through which we define the following conformity score:
\begin{equation}
V(X_i, Y_i) = \inf\{\tilde{\alpha} \in (0, 1): Y_i \in \widehat{\mathcal{C}}_{\tilde{\alpha}}(X_i)\},
\label{eq4nonconform}
\end{equation}
for all $i \in \mathcal{D}_c$. The conformity score in (\ref{eq4nonconform}) checks for the smallest value of $\tilde{\alpha}$ for which the corresponding interval $\widehat{\mathcal{C}}_{\tilde{\alpha}}(X_i)$ in $\widehat{\mathcal{C}}$ covers the response $Y_i$. We compute the empirical quantile of conformity scores $Q_\mathcal{V}(1-\alpha)$ as in (\ref{eq7}), and construct a predictive interval for $X_{n+1}=x$ as:
\begin{equation}
\widehat{C}_{n}(x) \coloneqq \{\widehat{\mu}(x) \pm \widehat{Q}(\tilde{\alpha}^*, x)\},\,  \tilde{\alpha}^* = Q_\mathcal{V}(1-\alpha).
\label{eq4intervals}
\end{equation} 
We drop the dependence of $\widehat{C}_{n}$ on $\alpha$ to reduce notational clutter. Note that the procedure in (\ref{eq4intervals}) produces predictive intervals that vary across prediction~instances~since it picks an entire conditional quantile function from the nested set. The intervals in (\ref{eq4intervals})~still~follow~the CP construction: hence, they satisfy the following marginal coverage guarantee.

{\bf Proposition 1.} {\it Consider a sequence of plug-in estimates $\{\widehat{Q}(\tilde{\alpha}, .)\}_{\tilde{\alpha}}$ obtained from a sample $\mathcal{D}_{c,1}$, and the corresponding conformity scores $\mathcal{V}=\{V(X_i, Y_i): i \in \mathcal{D}_{c,2}\}$ obtained from another sample $\mathcal{D}_{c,2}$, where $\mathcal{D}_{c,1}$ and $\mathcal{D}_{c,2}$ are two disjoint subsets of $\mathcal{D}_{c}$. If $\{(X_i, Y_i): 1 \leq i \leq n+1\}$ are exchangeable, then the interval in (\ref{eq4intervals}) satisfies} 
\begin{equation}
\mathbb{P}(Y_{n+1} \in \widehat{C}_{n}(X_{n+1})) \geq 1 - \alpha. \nonumber
\end{equation}
Proof is given in Appendix A. Variants of this result appear in the literature \cite{lei2018distribution, gupta2021nested}. Proposition 1 indicates~that,~by~defining conformity scores over a sequence of bands rather than intervals, we can construct adaptive predictive intervals while retaining the marginal coverage guarantees of CP. 

{\bf Plug-in estimation via UQR.}

We use UQR \cite{firpo2009unconditional} to fit the nested sequence of plug-in estimates in (\ref{eq4nested}). In what follows, we explain how UQR works from a Taylor approximation perspective. UQR approximates the conditional quantiles of $Y\,|\,X=x$, $Q(\alpha, x)$, under the true distribution $P$ as the marginal quantile of $Y$, $Q(\alpha)$, under an alternative distribution $G_x$ that is ``localized'' around $x$. Since the quantile is a statistical functional of the underlying distribution, we can estimate the marginal quantile under $G_{x}$ given the marginal quantile under $P$ using a von Mises linear approximation (VOM), i.e., a distributional analog of Taylor series of the following form \cite{fernholz2012mises}:
\begin{equation}
Q_{G_{x}}(\alpha) \approx Q_{P}(\alpha) + \int \mbox{IF}(y; Q(\alpha), P) \cdot dG_{x}(y),
\label{eqRIF16}
\end{equation}
where $\mbox{IF}$ is the influence function of the functional $Q(\alpha)$ at $P$ for a given point $(x, y)$ in the direction of the localized distribution $G_{x}$, which is defined as follows: 
\begin{equation}
\mbox{IF}(y; Q(\alpha), P) = \lim_{\epsilon \to 0} \frac{Q_{P^y}(\alpha)-Q_P(\alpha)}{\epsilon},
\label{eqRIF17}
\end{equation}
where $P^y = (1-\epsilon) P + \epsilon\, \delta_{y}$.
The influence function of the quantile measures the contribution of the outcome value $y$ on the marginal quantile statistic $Q_P(\alpha)$. By weighting the contributions of observations sampled from $P$ using the localized density $G_{x}$ as in (\ref{eqRIF16}), we obtain a first-order approximation of the marginal quantile functional under $G_{x}$. The influence function of the quantile functional is:
\begin{equation}
\mbox{IF}(y; Q(\alpha), P) = \frac{\alpha - \boldsymbol{1}\{y \leq Q_P(\alpha)\}}{f_Y(Q_P(\alpha))}.
\label{eqRIF18}
\end{equation}
Here, $f_Y(.)$ is the (one-dimensional) marginal density of $Y$. The derivation of the formula in (\ref{eqRIF18}) is standard, and is provided in Appendix B for completeness. The {\it re-centered} influence function (RIF) is defined as: 
\begin{equation}
\mbox{RIF}(y; Q(\alpha), P) = Q_P(\alpha) + \mbox{IF}(y; Q(\alpha), P),
\label{eqRIF18x}
\end{equation}
UQR involves regressing $\mbox{RIF}$ over $X$ to obtain~a~model~for $\mathbb{E}[\,\mbox{RIF}(Y; Q(\alpha), P)\,|\,X=x\,]$. Note that the influence function in (\ref{eqRIF18}) is a dichotomous variable as $\boldsymbol{1}\{y \leq Q_P(\alpha)\}$ is the only term that changes across covariates. Thus, UQR involves fitting a one-dimensional density estimate for $f_Y$ and a binary classifier for the dichotomous variable. UQR is typically used to study the effect of changing the covariate distribution on the marginal quantiles of outcomes, e.g., the effect of unionization on wages \cite{porter2015quantile}.  In our setup, we use (\ref{eqRIF16}) to obtain a plug-in estimate for the predictive band at the test point $x$ as follows:
\begin{equation}
Q_{G_{x}}(\alpha) \approx \int \mathbb{E}[\,\mbox{RIF}(Y; Q(\alpha), P)\,|\,X=x\,] \cdot dG_{x}.
\label{eqRIF19apx}
\end{equation} 

{\bf Constructing the nested sequence using UQR.}

The approximation in (\ref{eqRIF19apx}) inspires a simple regression procedure for constructing the nested sequence in (\ref{eq4nested}) while avoiding quantile crossing. Let $\mbox{RIF}_\alpha(X_i)$ be the RIF of the level-$\alpha$ quantile associated with the $i$-th data point. For a test point $X_{n+1}=x$, we can predict the value of $Q_{G_{x}}(\alpha)$ by fitting an ML model on the data set $\{(X_i, \mbox{RIF}_\alpha(X_i))\}_{i \in \mathcal{D}_c}$. Let the RIF values predicted by the ML model be $\widehat{\mbox{RIF}}_\alpha(x)$. Then, by repeating this process $K > 0$ times for all values of $\tilde{\alpha}$ in ${\bf \tilde{\alpha}} = [1/K, \ldots, (K-1)/K]$, we can construct the nested sequence as follows:\footnote{We take the absolute value of $\widehat{\mbox{RIF}}$ to account for erroneously negative predictions.} 
\begin{equation}
\widehat{\mathcal{C}}(x) = \{\widehat{\mathcal{C}}_k(x) = \widehat{\mu}(x) \pm |\widehat{\mbox{RIF}}_{\tilde{\alpha}_k}(x)|\}_{k=1}^{K-1}.
\label{eqRIF19xx}
\end{equation}  
where $\tilde{\alpha}_k=1/k$. Here, the RIF is defined with respect to quantiles of the model residual $E=|\,\widehat{\mu}(X)-Y|$ rather than the outcome $Y$. Note that, combining (\ref{eqRIF18}) and (\ref{eqRIF18x}), the RIF for the level $\tilde{\alpha}_k$ quantile can be written as:
\begin{equation}
\mbox{RIF}_{\tilde{\alpha}_k}(x) = Q_P(\tilde{\alpha}_k) + \frac{\tilde{\alpha}_k - \boldsymbol{1}\{e \leq Q_P(\tilde{\alpha}_k)\}}{f_E(Q_P(\tilde{\alpha}_k))}.
\label{eqRIF19secnd}
\end{equation}  
The 1-D density $f_E(.)$ can be estimated using kernel density estimation (KDE), and $Q_P(\alpha)$ can be estimated as the empirical level-$\tilde{\alpha}_k$ quantile of the residuals. The only term that we need to predict for each test point is $\boldsymbol{1}\{e \leq Q_P(\tilde{\alpha}_k)\}, \forall k \in \{1,\ldots, K-1\}$. This can be achieved with a single ML model as follows: for each data point $(X_i, E_i)$, define a target $k_i^* \coloneqq \min k,$ s.t. $E_i \leq \widehat{Q}_P(\tilde{\alpha}_k)$, then fit a model $g_\theta(.)$ (e.g., a regression model or a multi-class classifier) on the data set $\{(X_i, k^*_i)\}_i$. For a new test point $X_{n+1}=x$, we predict the RIF at $X_{n+1}=x$ by plugging in the predictions of $g_\theta$ into (\ref{eqRIF19secnd}) as follows:
\begin{equation}
\widehat{\mbox{RIF}}_{\tilde{\alpha}_k}(x) = \widehat{Q}_P(\tilde{\alpha}_k) + \frac{\tilde{\alpha}_k - \boldsymbol{1}\{g_\theta(x) \leq k\}}{\widehat{f}_E(\widehat{Q}_P(\tilde{\alpha}_k))},
\label{eqRIF19third}
\end{equation}  
$\forall k \in \{1, \ldots, K-1\}$. The nested set in (\ref{eqRIF19xx}) can thus be constructed using the $K-1$ predictions in (\ref{eqRIF19third}). Note that for any $k < k^\prime$, it is sufficient that $\partial \widehat{f}_E(\widehat{F}_E^{-1}(\alpha))/\partial \alpha <0$ for the monotonicity of the nested intervals to be preserved, i.e., $\widehat{\mbox{RIF}}_{\tilde{\alpha}_k}(x) < \widehat{\mbox{RIF}}_{\tilde{\alpha}_{k^\prime}}(x)$. This condition is met by various typical probability distributions (i.e., the exponential distribution). When this condition is violated for any two consecutive intervals in our empirical estimate, we replace $\widehat{f}_E(\widehat{Q}_P(\tilde{\alpha}_k))$ with $\widehat{f}_E(\widehat{Q}_P(\tilde{\alpha}_{k-1}))$ to enforce monotonicity.

Alternative approaches to constructing the set $\{\widehat{Q}(\alpha, .)\}_{\alpha}$ include fitting $K$ independent quantile regression models or a single distributional model (e.g., a Bayesian nonparameteric regression \cite{chipman2010bart}). The RIF-based construction of the nested set $\{\widehat{Q}(\alpha, .)\}_{\alpha}$ is more computationally and statistically efficient than either approach as it requires training a single ML model with labels that condense information about the conditional quantiles at levels $[1/K, \ldots, (K-1)/K]$. 

\begin{algorithm}[h]
    \SetKwInOut{Input}{Input}
    \SetKwInOut{Output}{Output}
    \vspace{.05in}	
    {\color{black}
    \textbf{CUQR} \mbox{\small $(\{(X_i, Y_i)\}_{i=1}^n, (X_{n+1}, Y_{n+1}), \alpha, \mu, G, K)$}\;
    \vspace{.05in} 
    \Input{\, Data set \mbox{\small $\{(X_i, Y_i)\}_{i=1}^n$}, test covariate \mbox{\small $X_{n+1}$}, model \mbox{\small $\mu$}, parameters $\alpha$, $G$ and $K$.}
    \vspace{.05in}	
    \Output{\, \mbox{\small $\widehat{C}_n(X_{n+1})$} and \mbox{\small $\widehat{\mathcal{S}}_n(X_{n+1})$}}}
    -------------------------------------------------------------
    \vspace{.05in}
	{\color{black}
	{\it Split the data set into a proper training set \mbox{\small $\mathcal{D}_{t}$} and 2 disjoint calibration~sets~\mbox{\small $\mathcal{D}_{c} = \mathcal{D}_{c,1} \cup \mathcal{D}_{c,2}$}}\;
	\vspace{.05in}
	\textbf{Using the training set \mbox{\small $\mathcal{D}_{t}$}, do the following:}\;
	\vspace{.03in}
	Fit the predictive model \mbox{\small $\widehat{\mu}: \mathcal{X}\to \mathbb{R}$} using \mbox{\small $\mathcal{D}_{t}$}\; 
	\vspace{.03in}
	Partition \mbox{\small $\mathcal{X}$} into relevance subgroups \mbox{\small $\{\widehat{\mathcal{S}}_g\}_{g=1}^G$}\; 
	\vspace{.1in}	
	\textbf{Using the calibration set \mbox{\small $\mathcal{D}_{c,1}$}, do the following:}\;
	\vspace{.03in}
	Fit \mbox{\footnotesize $\widehat{f}_E(.)$}, \mbox{\footnotesize $\widehat{Q}_P(\tilde{\alpha}_k)$} \& $g_\theta$. Plug the estimates in (\ref{eqRIF19third}) \; 
	\vspace{.03in}	
	Construct a nested sequence of intervals \mbox{\small $\widehat{\mathcal{C}}(x) = \{\widehat{\mathcal{C}}_{k}(x) = \widehat{\mu}(x) \pm |\widehat{\mbox{RIF}}_{\tilde{\alpha}_k}(x)|\}_{k=1}^K$}\; 
	\vspace{.1in}	
	\textbf{Using the calibration set \mbox{\small $\mathcal{D}_{c,2}$}, do the following:}\;
	\vspace{.03in}
	For all calibration data within each subgroup in \mbox{\small $\{\widehat{\mathcal{S}}_g\}_{g=1}^G$}, compute the conformity~scores~in~(\ref{eq4nonconform})\;
	\vspace{.03in}
	For each subgroup, select the tightest band in \mbox{\small $\widehat{\mathcal{C}}$} that achieves the target coverage as in (\ref{eq4intervals}). Let \mbox{\small $\widehat{\mathcal{C}}_{k(g)}(x)$} be the band selected for the $g$-th subgroup \;
	\vspace{.1in}	
	\textbf{For a new test covariate $X_{n+1}$, do the following:}\;
	\vspace{.03in}
	Identify subgroup \mbox{\small $g_{n+1}$} for which \mbox{\small $X_{n+1} \in \widehat{\mathcal{S}}_{g_{n+1}}$}\;
	\vspace{.03in}
	\mbox{\small $\widehat{C}_n(X_{n+1}) \gets \widehat{\mathcal{C}}_{k(g_{n+1})}(X_{n+1})$}, \mbox{\small $\widehat{\mathcal{S}}_n(X_{n+1}) \gets \widehat{\mathcal{S}}_{g_{n+1}}$}\;
	\vspace{.05in}	
	{\bf Return} $\widehat{C}_n(X_{n+1})$ and $\widehat{\mathcal{S}}_n(X_{n+1})$.
	\vspace{.05in}	}
    \caption{\textbf{Conformalized UQR}}
\end{algorithm}

\subsection{Step 2: Conformalizing UQR within subgroups} 
\label{sec32}
In Section \ref{sec31}, we developed a procedure that fulfills the adaptivity requirement while retaining the marginal validity of the non-adaptive CP method (Proposition 1). These marginal guarantees, however, do not meet the criteria for being transparent as they do not reflect the accuracy of the issued intervals for any given prediction. To provide more transparent guarantees, the~second~step~selects an interval from the nested set by applying the conformal procedure within local regions in covariate-space as follows:
\begin{itemize}
\item Partition the covariate space into $G$ subsets~$\{\widehat{\mathcal{S}}_g\}_{g=1}^G$.
\item Apply the conformal procedure in Section \ref{sec31} locally within each subgroup $g$ by picking a subgroup-specific band $\widehat{\mathcal{C}}_{k(g)}(x)$ from the set $\{\widehat{\mathcal{C}}_k(x)\}^{K-1}_{k=1}$. 
\item For a test point $X_{n+1}=x$, identify the relevance subgroup $g_{n+1}$ in which it belongs and report the subgroup \mbox{\small $\widehat{\mathcal{S}}_{g_{n+1}}$} and the corresponding interval \mbox{\small $\widehat{\mathcal{C}}_{k(g_{n+1})}(X_{n+1})$}.
\end{itemize} 
The steps involved in our conformalized UQR (CUQR) procedure are given in Algorithm 1. The procedure reports both a predictive interval that is specific to each instance, and a subgroup for which a desired level of accuracy is achieved. The relevance subgroups can either be learned from training data (e.g., using a clustering algorithm) or predetermined using application-specific knowledge (e.g., disease subtypes or protected attributes). The subgroup can be reported in the form of a cluster with a ``representative'' covariate value that is typical for this subgroup.~In~the~clinical example discussed earlier, our algorithm's output can represent the relevance subgroup in terms of a set of ``representative'' patients. 

Increasing the number of subgroups $G$ increases the granularity of the achieved average coverage at the cost of higher variance in realized coverage. $G$ can be set to larger values for larger sample sizes, e.g., $G=O(n)$, so that conditional coverage is achieved asymptotically. In the other extreme case when $G=1$, we recover the standard marginal coverage guarantee. In what follows, we state the theoretical coverage guarantee provided by the CUQR interval \mbox{\small $\widehat{C}_{n}(.)$}. 

{\bf Theorem 1.} {\it If \mbox{\small $\{(X_i, Y_i): i \in \mathcal{D}_c\} \cup \{(X_{n+1}, Y_{n+1})\}$} are exchangeable conditional on \mbox{\small $\{\mathcal{S}_g\}_{g=1}^G$}, and \mbox{\small $\mathbb{P}(\mathcal{S}_g) > \delta$},\mbox{$\forall g$} and some \mbox{\small $\delta >0$}, then for any $\alpha \in (0,1)$ and $n \in \mathbb{Z}_{+}$, we have} 
\begin{equation}
\mathbb{P}\left[\,Y_{n+1} \in \widehat{C}_{n}(X_{n+1}) \,\big|\, X_{n+1} \in \widehat{\mathcal{S}}_g, \mathcal{D}\,\right] \geq 1-\alpha-\frac{\lambda}{\sqrt{n_g}}, \nonumber
\label{eqthm1}
\end{equation}
\mbox{\small $\forall 1 \leq g \leq G$}, with probability at least \mbox{\small $1-2\exp(-2\lambda^2)$}, for any \mbox{\small $\lambda \geq \sqrt{\log(2)/2}$}. Theorem 1 states that coverage holds with high probability conditional on each relevance subgroup $g$ with a slack $\lambda/\sqrt{n_g}$, where $n_g$ is the number of calibration points in $\widehat{S}_g$. Thus, a PAC guarantee on coverage can be achieved within each subgroup by selecting the predictive band with an empirical coverage of $(1-\alpha) + \lambda/\sqrt{n_g}$. This suggests a trade off between the {\bf transparency} and {\bf efficiency} of the issued intervals---increasing the number of subgroups $G$ means that coverage will hold almost surely for more granular subsets of the input space, i.e., more transparency. This comes at the cost of longer intervals since $n_g$ will be smaller as $G$ increases. The bound in Theorem 1 follows from an application of the Dvoretzky-Kiefer-Wolfowitz inequality to the empirical CDF of the conformity scores \cite{massart1990tight}. The full proof is provided in Appendix C.

\section{Related work}
\label{Sec4}
{\bf Distributional regression} models that~directly~estimate~the conditional density $P_{Y|X=x}$ provide adaptive estimates of uncertainty. Broad classes of methods fall under this category; a non-exhaustive list includes: Bayesian non-parametric regression (e.g., using Gaussan processes \cite{sniekers2015adaptive} or regression trees \cite{chipman2010bart}), Bayesian neural nets \cite{kingma2015variational, hernandez2015probabilistic, ritter2018scalable}, and deep ensembles \cite{lakshminarayanan2017simple, fort2019deep}. Many of these models can provide accurate pragmatic estimates of predictive variance, but without the finite-sample coverage guarantees enabled by CP. The achieved coverage of distributional regression can be very sensitive to modeling choices (e.g., hyper-parameters or prior distributions). For instance, in Bayesian regression, the frequentist coverage achieved by posterior credible intervals with exact inference depends on the choice of the prior \cite{bayarri2004interplay}. With the more commonly used approximate inference methods (e.g., dropout or variational inference~\cite{gal2016dropout}), the induced posterior distributions may not concentrate asymptotically, resulting in poor coverage behavior \cite{osband2016risk, hron2017variational}. Distributional models are often used in conjunction with CP approaches to satisfy finite-sample coverage while maintaining adaptivity \cite{chernozhukov2021distributional}. 

{\bf Conformal prediction} in its most basic form achieves finite-sample marginal coverage at the expense of adaptivity \cite{vovk1999machine, vovk2005algorithmic, papadopoulos2008inductive}. Various approaches to CP-based adaptive predictive inference have been recently proposed \cite{lei2018distribution, romano2019conformalized, guan2019conformal,feldman2021improving,sesia2021conformal,gupta2021nested,guan2021localized,foygel2021limits, 
bai2022efficient}. The idea of ``conformalizing'' a plug-in estimate of the conditional quantile function originated in \cite{romano2019conformalized}. In this work, a single quantile regression model \mbox{\small $\widehat{Q}(\alpha, x)$} is fit to the training data, and a conformity score that measures the accuracy of \mbox{\small $\widehat{Q}(\alpha, x)$} is used to derive an adjustment for these intervals. Conformalized quantile regression provides marginal coverage guarantees, but its empirical conditional coverage depends on the quality of the underlying quantile regression model. Refinements of this approach were later proposed through two different lines of work: the first uses a re-weighting technique to ``localize'' CP at new test points \cite{guan2019conformal, guan2021localized}, and the second conformlizes a distributional regression model from which conditional quantiles can be derived \cite{sesia2021conformal, chernozhukov2021distributional}. In both lines of work, conditional validity is achieved in an asymptotic sense.

Our work holds subtle connections to these two approaches. In terms of the construction of conformity scores, \cite{sesia2021conformal, chernozhukov2021distributional} define the scores based on conditional ranks rather than error residuals---our procedure constructs predictive intervals by selecting among ``candidate'' estimates of conditional quantiles generated by varying the quantile level $\alpha$. This is equivalent to constructing the predictive intervals by selecting among estimates of conditional ranks of the full conditional distribution $P_{Y|X}$, but unlike the procedures in \cite{sesia2021conformal, chernozhukov2021distributional}, ours does not require access to consistent estimates of $P_{Y|X}$. Similar to the localized CP methods in \cite{guan2019conformal, guan2021localized}, our procedure is effectively a localized version of the marginal CP method. But unlike localized CP (LCP), our procedure can utilize any ML model for obtaining the localized quantile estimates (i.e., Equation (\ref{eqRIF19third})), whereas LCP is limited to re-weighting estimators (e.g., based on Nadarya-Watson kernel). Additionally, because our procedure selects among multiple quantile functions within each subgroup, it can achieve finite-sample (rather than asymptotic) coverage within local regions. The nested construction of our plug-in estimates falls within the general~nested CP formulation developed in \cite{gupta2021nested}. Unlike the formulation in this work, we construct our nested sets by parametrizing a functional form for the predictive band rather than directly parametrizing intervals, which enables selecting a different interval for each instance within a subgroup.

Re-centered influence functions (RIF) are typically used as targets for unconditional quantile regression models, a common modeling tool in econometric studies \cite{firpo2009unconditional}. To the best of our knowledge, RIF- based regression has not been operationalized  as a conformity score prior to this work.

\section{Experiments}
\label{Sec5}
We compare CUQR with various conformal and quantile regression baselines across multiple benchmark data~sets.~We start by describing our experimental setup below.

{\bf Baselines.} We consider standard split {\it conformal prediction} (CP), the {\it locally adaptive CP} (LACP) method in \cite{papadopoulos2011regression}, {\it conformalized quantile regression (CQR)} \cite{romano2019conformalized}, and {\it conformal conditional histograms} (CCH)  \cite{sesia2021conformal}. We also consider two variants of the standard {\it quantile regression} (QR) model for estimating conditional quantiles: QR with an underlying random forest model (QR-RF), and QR implemented using a neural network (QR-NN). We consider two ablated versions of CUQR that apply a conformalization procedure within the same relevance subgroups of CUQR. The first baseline, dubbed CQ, constructs the nested set $\mathcal{C}(.)$ using the empirical quantiles of residuals within the relevance subgroups, assigning the same intervals to all units within a subgroup without using the UQR-based plug-in estimates. The second baseline, which we call CQR-S,~applies~the~adaptive CQR method \cite{romano2019conformalized} within the relevance subgroups.

Finally, we consider two variants of our method: CUQR which applies conformalization based on the empirical $(1-\alpha)$-th quantiles, and CUQR-PAC, which corrects for the slack term in Theorem 1 to provide a high probability (PAC-style) coverage per subgroup. We select the value of $\lambda$ in Theorem 1 so that the probability that coverage holds (conditional on training data) is 90\%, i.e., $1-2\exp(-2\lambda^2) = 0.9$. We implement our method using an XGBoost regression model for $g_\theta$. In all experiments, we create the relevance subgroups using the $K$-means clustering algorithms. Further experimental details are provided in Appendix D.

{\bf Evaluation metrics.} We evaluate all baselines with respect to their achieved marginal coverage $C_{av}$, efficiency quantified via average interval length $L_{av}$ and subgroup-level coverage denoted as \mbox{\small $C_{av}(\widehat{\mathcal{S}}_g)$} for all subgroups \mbox{\small $\{\widehat{\mathcal{S}}_g\}_{g=1}^G$}. We also evaluate the worst-case subgroup-level coverage, defined as $C^{w.c.}_{G} = \min_g C_{av}(\widehat{S}_g)$. All metrics are evaluated on testing data and averaged over 10 runs. Unless otherwise stated, we set the target coverage level to $1-\alpha=0.9$.

{\bf Data sets.} We evaluate all baselines on {\bf 9 benchmark data sets} that are commonly used to evaluate CP methods: MEPS-19, MEPS-20, MEPS-21, Facebook-1, Facebook-2, Bio, Kin8nm, Naval, and Blog \cite{feldman2021improving, romano2019conformalized, feldman2021improving, chung2021beyond}.~Due~to~space~limitations, we highlight results for four data sets (MEPS-19, Facebook-1, Blog and Kin8nm) in this Section and defer further results to the Appendix. Details of all~data~sets~are provided in the Appendix. For each run, we randomly split each data set into disjoint training $\mathcal{D}_t$ (42.5$\%$), calibration $\mathcal{D}_c$ (42.5$\%$) and testing $\mathcal{D}_{test}$ (15$\%$) samples. For the LACP, CCH, CQR, CQR-S, CQ and CUQR baselines, we further split the calibration set $\mathcal{D}_c$ in half to obtain plug-in estimates and conformity scores from different splits.~In~all~experiments, we fit a Gradient Boosting regression model $\widehat{\mu}$ using the training set $\mathcal{D}_t$ and apply the predictive inference baselines on top of the predictions issued by the model $\widehat{\mu}$. 

\begin{table*}[t]
\centering
{\footnotesize
\begin{tabular}{llllllllllllllll}
\toprule
\toprule
       & \multicolumn{3}{c}{\textbf{MEPS-19}} & \multicolumn{3}{c}{\textbf{Facebook-1}} & \multicolumn{3}{c}{\textbf{Blog}} & \multicolumn{3}{c}{\textbf{Kin8nm}}  \\ \midrule 
               &  $C_{av}$  & $L_{av}$     & $C^{w.c.}_{G}$  & $C_{av}$     & $L_{av}$   & $C^{w.c.}_{G}$    & $C_{av}$     & $L_{av}$  & $C^{w.c.}_{G}$ & $C_{av}$     & $L_{av}$     & $C^{w.c.}_{G}$ \\ \midrule
\textbf{QR methods} &&&&&&&&&&&&& \\
\quad QR-RF          & 0.90 & 1.00 & 0.54 & 0.93 & 0.85 & 0.78 & 0.79 & 0.73 & 0.76 & 0.93 & 1.36 & 0.89  \\
\quad QR-NN          & 0.79 & 0.54 & 0.67 & 0.81 & 0.55 & 0.68 & 0.79 & 0.73 & 0.76 & 0.79 & 0.94 & 0.74 \\
\midrule
\textbf{CP methods} &&&&&&&&&&&&& \\
\quad CP             & 0.89 & 1.28 & 0.19 & 0.90 & 1.39 & 0.72 & 0.89 & 1.89 & 0.57 & 0.90 & 2.17 & 0.83  \\
\quad LACP           & 0.89 & 0.61 & 0.20 & 0.90 & 0.69 & 0.76 & 0.89 & 1.06 & 0.63 & 0.90 & 1.09 & 0.84  \\
\quad CQR            & 0.89 & 1.12 & 0.46 & 0.90 & 0.83 & 0.77 & 0.90 & 1.34 & 0.82 & 0.90 & 1.33 & 0.85  \\
\quad CCH            & 0.96 & 5.37 & 0.79 & 0.89 & 0.72 & 0.65 & 0.98 & 5.58 & 0.96 & 0.89 & 1.14 & 0.86  \\ 
\quad CQ             & 0.87 & 2.02 & 0.76 & 0.89 & 1.34 & 0.79 & 0.87 & 1.81 & 0.76 & 0.89 & 2.16 & 0.85 \\
\quad CQR-S          & 0.89 & 1.54 & 0.67 & 0.90 & 0.77 & 0.87 & 0.90 & 1.37 & 0.80 & 0.90 & 1.33 & 0.85 \\\midrule
\quad {\bf CUQR}     & 0.89 & 1.25 & 0.73 & 0.90 & 1.36 & 0.87 & 0.87 & 1.82 & 0.67 & 0.89 & 2.19 & 0.85 \\
\quad {\bf CUQR-PAC} & 0.89 & 2.90 & 0.88 & 0.92 & 1.61 & 0.90 & 0.90 &	2.23 & 0.82 & 0.96 & 3.27 & 0.93 \\
\bottomrule
\bottomrule
\end{tabular}}
\vspace{.05in}
\caption{\footnotesize Marginal coverage, efficiency and conditional coverage of all baselines on benchmark data sets.}
  \rule{\linewidth}{.75pt}   
  \vspace{.05in} 
\end{table*}

\subsubsection*{Results}

{\bf Evaluating transparency.} All baselines are calibrated to have a target marginal coverage of 90$\%$, but what does this notion of coverage mean to individual users of the model? In this experiment, we assess the transparency of different baselines by evaluating the worst-case~conditional~coverage within typical ``subgroups'' of individuals or prediction instances. We use $K$-means clustering to~identify~$G=10$~relevance subgroups using training samples.~Note~that~the~subgroups are not arbitrary---they represent a clustering of the population into ``typical'' subgroups of similar individuals.

In Table 1, we show the marginal coverage and average lengths of predictive intervals for all baselines across the three data sets. First, we observe that while all conformal methods achieve the target (marginal) coverage levels, the coverage of the QR baselines vary depending on the underlying model specification, which is expected as these baselines do not provide any coverage guarantees. On the contrary, all CP-based methods achieve their promised model-agnostic coverage guarantees, but how well do the different CP variants perform when examined on a subgroup level?  

As we can see in Table 1, CP baselines that only guarantee marginal coverage (CP and CQR) have a very poor worst-case coverage conditional on a subgroup, i.e., their declared guarantees do not reflect their performance among a large subgroup of ``similar'' individuals in a given population. Similary, CP methods with asymptotic conditional coverage guarantees can exhibit severe under-coverage in finite samples (LACP), or maintain reasonable conditional coverage but with poor efficiency (CCH), which highlights the importance of controlling for finite-sample conditional coverage. While CUQR achieves subgroup-specific coverage marginally, the worst-case subgroup-level~coverage~can be significantly lower than the desired target coverage for some data sets (e.g., MEPS-19 and Blog). The CUQR-PAC variant of our method guarantees that~coverage~holds~within each subgroup with high probability. This comes at the cost of efficiency, i.e., wider predictive intervals.
 
\begin{figure}[h]
    \centering
    \includegraphics[scale=0.325]{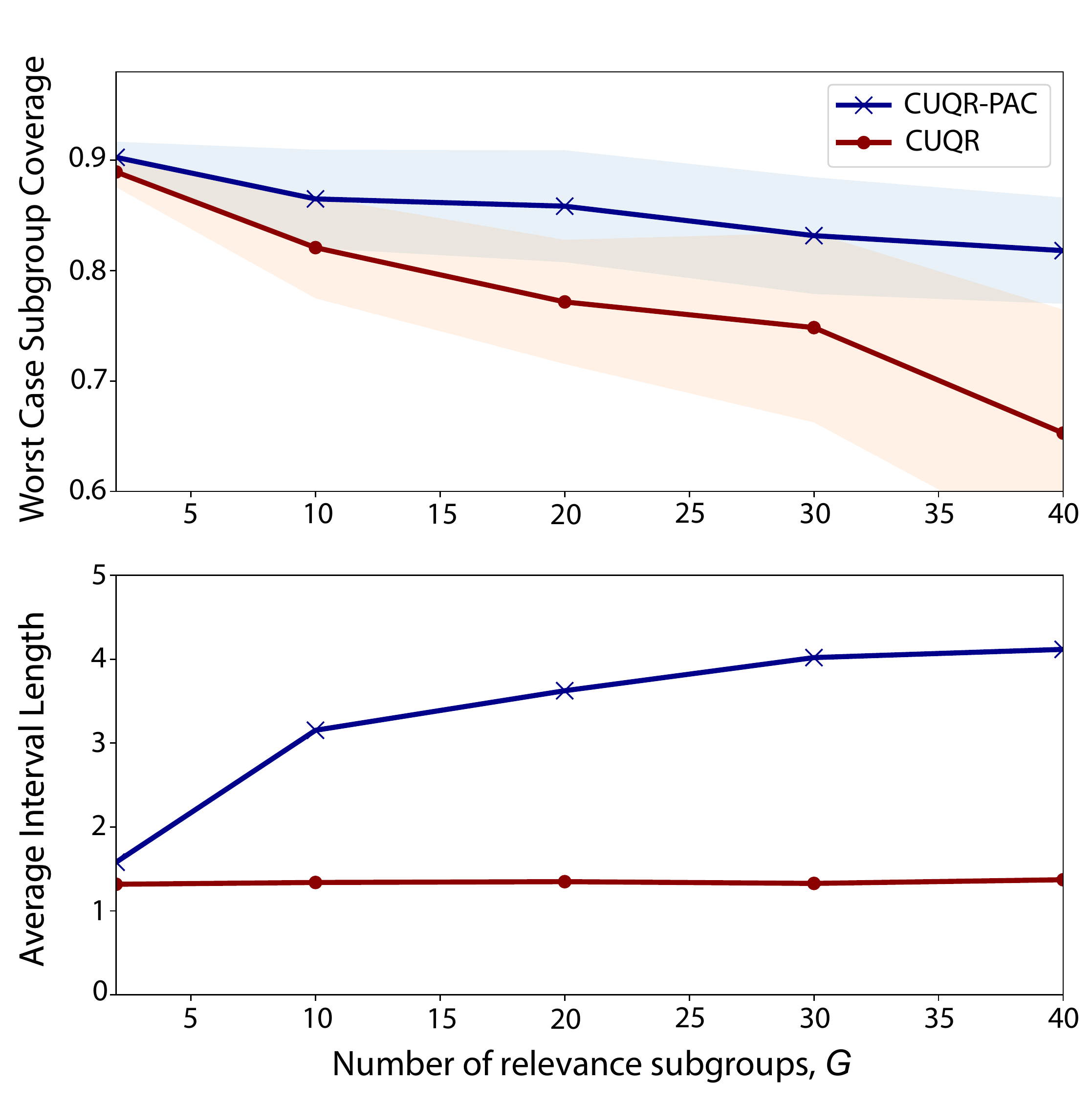}
    \caption{\footnotesize Impact of the number of subgroups $G$ on coverage.}
    \label{fig:my_label0} 
    \rule{\linewidth}{.75pt} 
\end{figure}

The CQ variant of our method, which picks a fixed interval per subgroup rather than a full RIF-based predictive band, achieves better worst-case coverage at the expense of within-subgroup adaptivity and average efficiency. Because the subgroup-specific coverage guarantees for CUQR hold on average with respect to the randomness of calibration data, the variance of the empirically achieved coverage on test data increases as the number of subgroups $G$ increases (i.e., smaller calibration sample per~subgroup).~Consequently,~the worst case subgroup-level coverage achieved by CUQR decreases (in expectation) as the number of relevance subgroups increases. To manage the trade off between transparency and efficiency, the CUQR-PAC variant of our method accounts for the variance of realized coverage by inflating the predictive intervals to achieve an empirical coverage of $(1-\alpha)+\lambda/\sqrt{n_g}$ (See Figure S1 \ref{fig:my_label0}).

\begin{figure*}[t]
    \centering
    \includegraphics[scale=0.45]{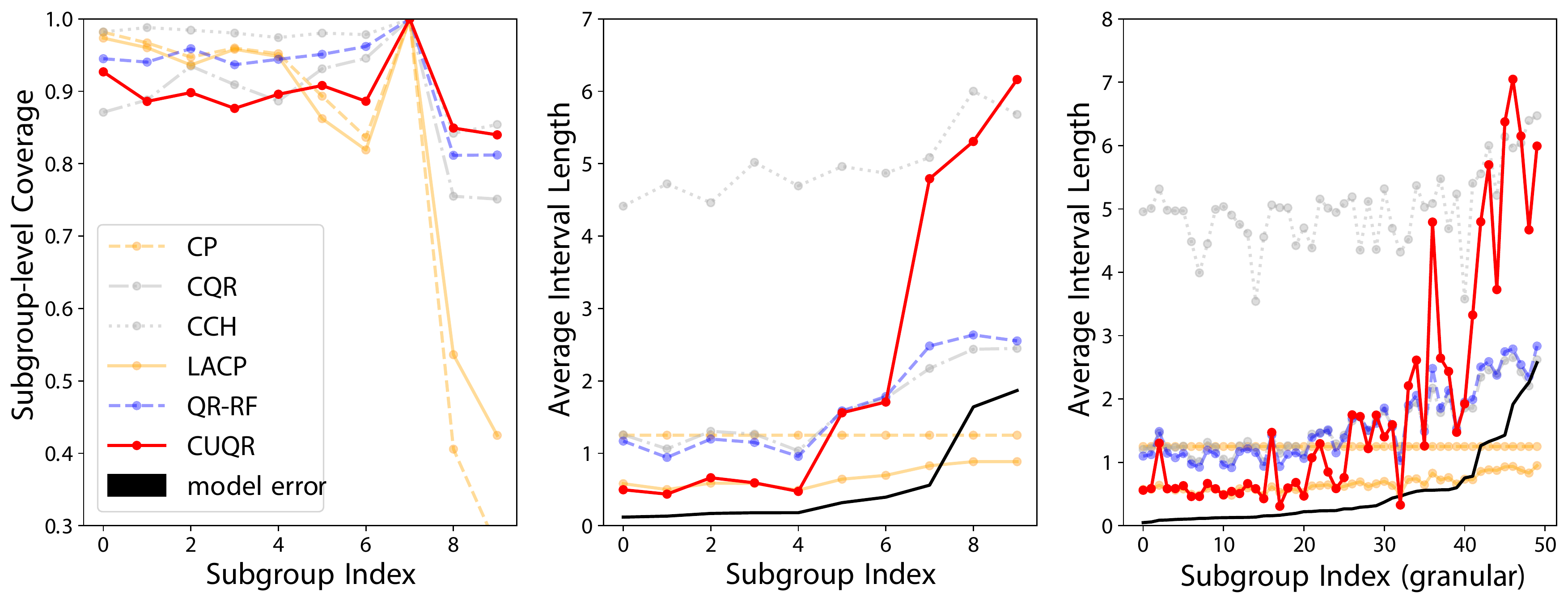}
    \caption{\footnotesize Adaptivity of predictive inference baselines on the {\bf MEPS-19} dataset.}
    \label{fig:my_label}
    \rule{\linewidth}{.75pt}   
\end{figure*}

{\bf Evaluating adaptivity.} Next, we assess the extent to which baselines are adaptive, i.e., the lengths of their intervals vary according to the true uncertainty of the base~model~$\widehat{\mu}$. Among the data sets under study,~the~MEPS-19~data~exhibited significant heteroscedasticity, i.e., the average error of the predictive model varies significantly~across~the~subgroups. Hence, we expect the average lengths of intervals issued by adaptive procedures to be greater for subgroups where the model errors are high. In Figure \ref{fig:my_label}, we plot the achieved subgroup-level coverage (left) and the corresponding average interval length per subgroup (middle) in the {\bf MEPS-19} dataset. In both Figures, the subgroup indexes on the $x$-axis are ordered ascendingly according to the model's subgroup-level average error on testing data (i.e., larger indexes correspond to higher model uncertainty). (In Figure \ref{fig:my_label}, we exclude baselines that were under-performing to avoid clutter.) As we can see, CUQR maintains the target coverage approximately for all subgroups, and adjusts the lengths of its issued intervals withing each subgroup according to the model uncertainty. On the contrary, competing baselines either fail to recognize the varying uncertainty~across~subgroups (LACP and CP), or do not adequately~adapt~the~interval length to maintain target coverage (QR-RF and CQR).

Finally, to evaluate the adaptivity of CUQR beyond the subgroups on which it was calibrated, we run a $K$-means clustering algorithm with a different random seed and different number of clusters $G=50$, and order the subgroup indexes ascendingly according to the base model's uncertainty as before. We evaluate the average interval lengths of CUQR (previously fitted on the $G=10$ subgroups), along with the other baselines, on the new and more granular 50 subgroups. As we can see in Figure \ref{fig:my_label} (right), CUQR outperforms other baselines in adapting its intervals to subgroup-level uncertainty, indicating better conditional adaptivity properties beyond what is implied by the theoretical guarantees.  

\section{Conclusion}
In this paper, we developed a conformal prediction method that adapts its issued intervals to the level of uncertainty in each prediction instance, while reporting a local region in covariate-space that contains the queried covariate instance and over which the procedure is guaranteed to be accurate on average. Our procedure partitions the covariate~space~into subgroups, and leverages the re-centered influence function of the quantile functional to construct a nested sequence of predictive bands, from which it selects one band per subgroup. By reporting instance-specific predictive intervals and subgroup-specific coverage guarantees to end-users, our method enables a more transparent approach to communicating uncertainty in the predictions of ML models. 


\bibliographystyle{unsrt}
\bibliography{uncertainty}

\newpage

\onecolumn
\section*{SUPPLEMENTARY MATERIAL}

\subsection*{Appendix A: Proof of Proposition 1}
{\bf Proposition 1.} {\it Consider a sequence of plug-in estimates $\{\widehat{Q}(\tilde{\alpha}, .)\}_{\tilde{\alpha}}$ obtained~from~a~calibration~sample $\mathcal{D}_{c,1}$, and the corresponding conformity scores $\mathcal{V}=\{V(X_i, Y_i): i \in \mathcal{D}_{c,2}\}$~obtained from another sample $\mathcal{D}_{c,2}$, where $\mathcal{D}_{c,1}$ and $\mathcal{D}_{c,2}$ are two disjoint subsets of $\mathcal{D}_{c}$. If $\{(X_i, Y_i): 1 \leq i \leq n+1\}$ are exchangeable, then the interval in (\ref{eq4intervals}) satisfies} 
\begin{equation}
\mathbb{P}(Y_{n+1} \in \widehat{C}_{n}(X_{n+1})) \geq 1 - \alpha. \nonumber
\end{equation}
{\it Proof.} The construction of the interval in (\ref{eq4intervals}) implies that, conditioned on $\mathcal{D}_{c,1}$, we have 
\begin{align}
Y_{n+1} \in \widehat{C}_n(X_{n+1}) \iff V(X_{n+1}, Y_{n+1}) \leq  Q_\mathcal{V}(1-\alpha),\nonumber
\end{align}
hence it follows that \mbox{\small $\mathbb{P}(Y_{n+1} \in \widehat{C}_n(X_{n+1})\,|\,\mathcal{D}_{c,1})) = \mathbb{P}(V(X_{n+1}, Y_{n+1} \leq  Q_\mathcal{V}(1-\alpha)\,|\,\mathcal{D}_{c,1}))$}. Since all training and calibration samples $(X_i, Y_i)$ are exchangeable, then conditioned on $\mathcal{D}_{c,1}$, the calibration conformity scores in $\mathcal{V}$ and the conformity score on the test~point~\mbox{\small $V(X_{n+1}, Y_{n+1})$}~are~exchangeable. By exchangeability of calibration and testing conformity scores, it follows from Lemma 2 in \cite{romano2019conformalized} that:
\begin{align}
\mathbb{P}(V(X_{n+1}, Y_{n+1}) \leq  Q_\mathcal{V}(1-\alpha)\,|\,\mathcal{D}_{c,1}) \geq 1-\alpha,\nonumber
\end{align}
which concludes the statement after marginalizing over the randomness of $\mathcal{D}_{c,1}$. \qed

\subsection*{Appendix B: Influence function of the quantile functional}
Recall that $Q(\alpha, x)$ is the level-$\alpha$ quantile of $Y|X=x$, i.e., 
\[Q(\alpha, x) = F^{-1}(\alpha) \coloneqq \inf\{y \in \mathbb{R}: F(y\,|\,X=x) \geq \alpha\},\] 
where $F(.)$ is the conditional cumulative density function (CDF), $F(y\,|\,X=x) \coloneqq \mathbb{P}(Y \leq y\,|\,X=x)$. Throughout this Section, we assume that $\alpha = F(F^{-1}(\alpha))$. Recall that the influence function of the functional $Q(\alpha)$ at the distribution $P$ for a given point $(x, y)$ in the direction of the localized distribution $G_{x}$ is defined as follows: 
\begin{equation}
\mbox{IF}(y; Q(\alpha), P) = \lim_{\varepsilon \to 0} \frac{Q_{P^y}(\alpha)-Q_P(\alpha)}{\varepsilon},
\label{eqRIF17xxx}
\end{equation}
where $P_\varepsilon = (1-\varepsilon) P + \epsilon\, \delta_{y}$. Let $F_\varepsilon = (1-\varepsilon)F + \varepsilon \cdot \delta_y$ be the CDF of the perturbed distribution, then we have 
\begin{align}
\alpha &= F_\varepsilon \circ F_\varepsilon^{-1}(\alpha) \nonumber \\
&= (1-\varepsilon) \cdot F(F^{-1}_\varepsilon(\alpha)) + \varepsilon \cdot \delta_y(F^{-1}_\varepsilon(\alpha)).
\label{eqRIF17yyy}
\end{align}
By differentiating both sides with respect to $\varepsilon$, we have the following
\begin{align}
\frac{\partial \alpha}{\partial \varepsilon} = 0 &= \frac{\partial}{\partial \varepsilon}\left(F_\varepsilon \circ F_\varepsilon^{-1}(\alpha)\right) \nonumber \\
&= \frac{\partial}{\partial \varepsilon}\left((1-\varepsilon) \cdot F(F^{-1}_\varepsilon(\alpha)) + \varepsilon \cdot \delta_y(F^{-1}_\varepsilon(\alpha))\right) \nonumber \\
&= -F(F_\varepsilon^{-1}(\alpha)) + (1-\varepsilon) f(F_\varepsilon^{-1}(\alpha)) \cdot \frac{\partial F_\varepsilon^{-1}(\alpha)}{\partial \varepsilon} + \delta_y(F_\varepsilon^{-1}(\alpha))  + \varepsilon \cdot \frac{\partial}{\partial \varepsilon} \delta_y(F_\varepsilon^{-1}(\alpha)).
\label{eqRIF17zzz}
\end{align}
Note that the influence function definition in (\ref{eqRIF17xxx}) can be rewritten as follows:
\begin{equation}
\mbox{IF}(y; Q(\alpha), P) = \lim_{\varepsilon \to 0} \frac{Q_{P^y}(\alpha)-Q_P(\alpha)}{\varepsilon} = \left(\frac{\partial}{\partial \varepsilon} F_\varepsilon^{-1}(\alpha)\right)_{\varepsilon=0}.
\label{eqRIF17sss}
\end{equation}
By setting $\varepsilon=0$ in (\ref{eqRIF17zzz}), we have the following: 
\begin{align}
0 &= \left(-F(F_\varepsilon^{-1}(\alpha)) + (1-\varepsilon) f(F_\varepsilon^{-1}(\alpha)) \cdot \frac{\partial F_\varepsilon^{-1}(\alpha)}{\partial \varepsilon} + \delta_y(F_\varepsilon^{-1}(\alpha))  + \varepsilon \cdot \frac{\partial}{\partial \varepsilon} \delta_y(F_\varepsilon^{-1}(\alpha))\right)_{\varepsilon=0} \nonumber \\
&= -F(F^{-1}(\alpha)) + f(F^{-1}(\alpha)) \cdot \left(\frac{\partial F^{-1}_\varepsilon(\alpha)}{\partial \varepsilon}\right)_{\varepsilon=0} + \delta_y(F^{-1}(\alpha)).
\label{eqRIF17zzz2}
\end{align}
By rearranging the terms in (\ref{eqRIF17zzz2}), we have
\begin{align}
\left(\frac{\partial F^{-1}_\varepsilon(\alpha)}{\partial \varepsilon}\right)_{\varepsilon=0} &= \frac{F(F^{-1}(\alpha)) - \delta_y(F^{-1}(\alpha))}{f(F^{-1}(\alpha))}.
\label{eqRIF17zzz3}
\end{align}
Hence, the influence function of the quantile is given by:
\begin{align}
\mbox{IF}(y; Q(\alpha), P) = \left(\frac{\partial F^{-1}_\varepsilon(\alpha)}{\partial \varepsilon}\right)_{\varepsilon=0} = \frac{\alpha - \boldsymbol{1}_{\{F^{-1}(\alpha) \geq y\}}}{f(F^{-1}(\alpha))}.
\label{eqRIF17zzz4}
\end{align}

\subsection*{Appendix C: Proof of Theorem 1}

We start by stating the Dvoretzky-Kiefer-Wolfowitz (DKW) inequality which will be used to prove Theorem 1. Let $\widehat{F}_n$ be the empirical distribution function for $n$ i.i.d random variables with distribution function $F$, then we have
\begin{align}
\mathbb{P}\left(\sup_x \, (\widehat{F}_n(x) - F(x)) > \frac{\lambda}{\sqrt{n}}\right) \leq \exp(-2\lambda^2),    
\end{align}
for any $\lambda$ such that $\exp(-2\lambda^2)\leq \frac{1}{2}$, or equivalently, $\lambda \geq \sqrt{\log(2)/2}$. The proof is provided in \cite{massart1990tight}.

{\bf Theorem 1.} {\it If $\{(X_i, Y_i): i \in \mathcal{D}_c\} \cup \{(X_{n+1}, Y_{n+1})\}$ are exchangeable conditional on $\{\mathcal{S}_g\}_{g=1}^G$, and $\mathbb{P}(\mathcal{S}_g) > \delta$, $\forall g$ and some $\delta >0$, then for any $\alpha \in (0,1)$ and $n \in \mathbb{Z}_{+}$, we have} 
\begin{equation}
\mathbb{P}\left[\,Y_{n+1} \in \widehat{C}_{n}(X_{n+1}) \,\big|\, X_{n+1} \in \widehat{\mathcal{S}}_g, \mathcal{D}\,\right] \geq 1-\alpha-\frac{\lambda}{\sqrt{n_g}}, \nonumber
\label{eqthm1}
\end{equation}
$\forall 1 \leq g \leq G$, with probability at least $1-2\exp(-2\lambda^2)$, for any $\lambda \geq \sqrt{\log(2)/2}$.

{\it Proof.} 
Fix a relevance subgroup $g \in \{1, \ldots, G\}$. By constructing the interval in (\ref{eq4intervals}) within $\mathcal{S}_g$, a coverage event is:
\begin{align}
Y_{n+1} \in \widehat{C}_n(X_{n+1})\,|\, X_{n+1} \in \widehat{\mathcal{S}}_g \iff V(X_{n+1}, Y_{n+1})\leq  Q_{\mathcal{V}_g}(1-\alpha)\,|\, X_{n+1} \in \widehat{\mathcal{S}}_g ,
\label{equivxx}
\end{align}
where \mbox{\small$\mathcal{V}_g = \{V(X_i, Y_i): i \in \mathcal{D}_c, X_i \in \widehat{S}_g\}$}. Since {$\{(X_i, Y_i): i \in \mathcal{D}_c\}$} are exchangeable conditional on \mbox{\small $\{\widehat{S}_{g^\prime}\}_{g^\prime=1}^G$}, then the sequence of conformity scores \mbox{\small $\mathcal{V}_g$} and \mbox{\small $V(X_{n+1}, Y_{n+1})$} are exchangeable conditional on \mbox{\small $\{\widehat{S}_{g^\prime}\}_{g^\prime=1}^G$} and the event \mbox{\small $X_{n+1} \in \widehat{\mathcal{S}}_g$}. Hence, it follows from Lemma 2 in \cite{romano2019conformalized} that:
\begin{align}
\mathbb{P}(V(X_{n+1}, Y_{n+1}) \leq  Q_{\mathcal{V}_g}(1-\alpha)\,|\,X_{n+1} \in \widehat{\mathcal{S}}_g) \geq 1-\alpha.\nonumber
\end{align}
Conditional on the training data $\mathcal{D}$, we can bound the probability of the coverage event by applying the DKW inequality to the empirical estimate of the CDF of the conformity scores as follows: 
\begin{align}
\mathbb{P}\left(\sup_{v} \,\, \left(\frac{1}{n_g}\sum_{i=1}^{n_g} \boldsymbol{1}_{\{V(X_{n+1}, Y_{n+1})\leq  v\}}-F(v)\right) > \frac{\lambda}{\sqrt{n_g}}\,\,\Bigg|\,\,X_{n+1} \in \widehat{\mathcal{S}}_g, \mathcal{D}\right) \leq \exp(-2\lambda^2), 
\label{eqbound}
\end{align}
hence it follows that
\begin{align}
\mathbb{P}\left(\frac{1}{n_g}\sum_{i=1}^{n_g} \boldsymbol{1}_{\{V(X_{n+1}, Y_{n+1})\leq  Q_{\mathcal{V}_g}(1-\alpha)\}}-(1-\alpha) > \frac{\lambda}{\sqrt{n_g}}\,\,\Bigg|\,\,X_{n+1} \in \widehat{\mathcal{S}}_g, \mathcal{D}\right) \leq \exp(-2\lambda^2), 
\label{eqbound}
\end{align}
$\forall \lambda \geq \sqrt{\log(2)/2}$. From the equivalence relation in (\ref{equivxx}), we can write the bound in (\ref{eqbound}) as follows: 
\begin{align}
\mathbb{P}\left(\frac{1}{n_g}\sum_{i=1}^{n_g} \boldsymbol{1}_{\{Y_{n+1} \in \widehat{C}_n(X_{n+1})\}} > 1-\alpha-\frac{\lambda}{\sqrt{n_g}}\,\,\Bigg|\,\,X_{n+1} \in \widehat{\mathcal{S}}_g, \mathcal{D}\right) \leq \exp(-2\lambda^2), 
\label{eqbound2}
\end{align}
which concludes the proof. \qed 

\newpage
\subsection*{Appendix D: Additional Experiments}

\begin{table}
\centering
{\footnotesize
\begin{tabular}{llllllllll}
\toprule
\toprule
                 & \multicolumn{3}{c}{\textbf{MEPS-19}} & \multicolumn{3}{c}{\textbf{MEPS-20}} & \multicolumn{3}{c}{\textbf{MEPS-21}} \\ \midrule 
               & $C_{av}$     & $L_{av}$     & $C^{w.c.}_{G}$  & $C_{av}$     & $L_{av}$   & $C^{w.c.}_{G}$    & $C_{av}$     & $L_{av}$     & $C^{w.c.}_{G}$ \\ \midrule
\textbf{QR methods} &&&&&&&&& \\
\quad QR-RF   & 0.90 & 1.00 & 0.54 & 0.91 & 1.06 & 0.55 & 0.93 & 1.29 & 0.65 \\
\quad QR-NN   & 0.79 & 0.54 & 0.67 & 0.81 & 0.56 & 0.64 & 0.79 & 0.53 & 0.64 \\
\midrule
\textbf{CP methods} &&&&&&&&& \\
\quad CP    & 0.89 & 1.28 & 0.19 & 0.90 & 1.24 & 0.15 & 0.90 & 1.29 & 0.16 \\
\quad LACP  & 0.89 & 0.61 & 0.20 & 0.89 & 0.59 & 0.20 & 0.90 & 0.62 & 0.26 \\
\quad CQR   & 0.89 & 1.12 & 0.46 & 0.89 & 1.05 & 0.50 & 0.89 & 1.27 & 0.62 \\
\quad CCH   & 0.96 & 5.37 & 0.79 & 0.97 & 5.35 & 0.75 & 0.97 & 5.35 & 0.78 \\ 
\quad CQ  & 0.87 & 2.02 & 0.76 & 0.88 & 2.01 & 0.87 & 0.88 & 2.10 & 0.88 \\\midrule
\quad CUQR & 0.89 & 1.25 & 0.73 & 0.89 & 1.21 & 0.76 & 0.90 & 1.26 & 0.81 \\
\toprule
\toprule
                 & \multicolumn{3}{c}{\textbf{Facebook-1}} & \multicolumn{3}{c}{\textbf{Facebook-2}} & \multicolumn{3}{c}{\textbf{Bio}} \\ \midrule 
               & $C_{av}$     & $L_{av}$     & $C^{w.c.}_{G}$  & $C_{av}$     & $L_{av}$   & $C^{w.c.}_{G}$    & $C_{av}$     & $L_{av}$     & $C^{w.c.}_{G}$ \\ \midrule
\textbf{QR methods} &&&&&&&&& \\
\quad QR-RF   & 0.93 & 0.85 & 0.78 & 0.92 & 0.81 &0.86  & 0.92 & 1.33 & 0.84 \\
\quad QR-NN  & 0.81 & 0.55 & 0.68 & 0.80 & 0.52& 0.64 & 0.81 & 0.94 & 0.56 \\
\midrule
\textbf{CP methods} &&&&&&&&& \\
\quad CP   & 0.90  & 1.39 & 0.72 & 0.90 & 1.39 & 0.82 & 0.90 & 2.42 & 0.85\\
\quad LACP  & 0.90 & 0.69 & 0.76& 0.90 & 0.69 & 0.84 & 0.90 & 1.15 & 0.83 \\
\quad CQR   & 0.90 & 0.83 & 0.77& 0.90 & 0.83 & 0.87 & 0.90 & 1.36 & 0.83\\
\quad CCH   & 0.89 & 0.72 & 0.65& 0.89 & 0.64 & 0.67 & 0.90 & 1.07 & 0.85\\ 
\quad CQ  & 0.89 & 1.34 & 0.79 & 0.89 & 1.35 & 0.87 & 0.90 & 2.41 & 0.80 \\\midrule
\quad CUQR  & 0.90  & 1.36  & 0.87 & 0.89 & 1.36 & 0.87 & 0.90 & 2.40 & 0.88 \\

\toprule
\toprule
                 & \multicolumn{3}{c}{\textbf{Kin8nm}} & \multicolumn{3}{c}{\textbf{Naval}} & \multicolumn{3}{c}{\textbf{Blog}} \\ \midrule 
               & $C_{av}$     & $L_{av}$     & $C^{w.c.}_{G}$  & $C_{av}$     & $L_{av}$   & $C^{w.c.}_{G}$    & $C_{av}$     & $L_{av}$     & $C^{w.c.}_{G}$ \\ \midrule
\textbf{QR methods} &&&&&&&&& \\
\quad QR-RF   & 0.93 & 1.36 & 0.89 & 0.90 & 0.63 & 0.87 & 0.79 & 0.73 & 0.76 \\
\quad QR-NN   & 0.79 & 0.94 & 0.74 & 0.78 & 0.55 & 0.73 & 0.79 & 0.73 & 0.76 \\
\midrule
\textbf{CP methods} &&&&&&&&& \\
\quad CP   & 0.90 & 2.17 & 0.83 & 0.89 & 1.31 & 0.78 & 0.89 & 1.89 & 0.57 \\
\quad LACP & 0.90 & 1.09 & 0.84 & 0.89 & 0.60 & 0.85 & 0.89 & 1.06 & 0.63 \\
\quad CQR  & 0.90 & 1.33 & 0.85 & 0.89 & 0.70 & 0.85 & 0.90 & 1.34 & 0.82 \\
\quad CCH  & 0.89 & 1.14 & 0.86 & 0.89 & 0.48 & 0.83 & 0.98 & 5.58 & 0.96 \\ 
\quad CQ  & 0.89 & 2.16 & 0.85 & 0.87 & 1.27 & 0.87 & 0.87 & 1.81 & 0.76 \\ \midrule
\quad CUQR  & 0.89 & 2.19 & 0.85 & 0.86 & 1.26 & 0.85 & 0.87 & 1.82 & 0.67 \\
\bottomrule
\bottomrule
\end{tabular}}
\vspace{.05in}
\caption{\footnotesize Marginal coverage, efficiency and conditional coverage of all baselines on benchmark data sets.}
  \vspace{-2mm} 
  \rule{\linewidth}{.75pt}   
\label{tab:sup_tablev2}
\end{table}

We test our method on 9 datasets: \textbf{MEPS-19}, \textbf{MEPS-20}, \textbf{MEPS-21}, \textbf{Facebook-1}, \textbf{Facebook-2}, \textbf{Bio}, \textbf{Kin8nm}, \textbf{Naval} and \textbf{Blog}. The MEPS (medical expenditure panel survey) datasets comprise surveys of families and individuals, their medical providers, and employers across the United States collected over three years \cite{romano2019conformalized, feldman2021improving, chung2021beyond}. The facebook datasets contain features extracted from facebook posts, and the task is to predict how many comments the post will receive. The difference between \textbf{Facebook-1} and \textbf{Facebook-2} are the features that are included. The \textbf{Bio} dataset consists of features describing the physiochemical properties of protein tertiary structure, including fractional areas of different regions of the protein, distances between particular amino acids, and other spacial distribution constraints. The task is to predict the size of the residue. The remaining data sets are available in the UCI repository.

\textbf{Evaluating Transparency} In Table \ref{tab:sup_tablev2}, we show the marginal coverage, worst-case subgroup coverage, and average interval length for all baselines as well as our CUQR approach on the three additional datasets. In general, we recapitulate the observations that we discuss in the main paper, though we will take care to provide additional analysis for some of the more competitive baselines. Regarding the QR-RF baseline, we find that although it is superior in terms of efficiency, it does not have theoretical guarantees in terms of marginal or conditional coverage. Indeed, the conditional coverage is worse than the worst-case conditional coverage achieved by CUQR in all three datasets. With respect to CQR, although it is competitive in the \textbf{Facebook-2} dataset, it does not have any theoretical guarantees for coverage at the subgroup level (only marginal coverage is guaranteed). Consequently, we see for \textbf{Facebook-1} and \textbf{Bio} inferior conditional coverage compared to CUQR.

\begin{figure}
    \centering
    \includegraphics[scale=0.38]{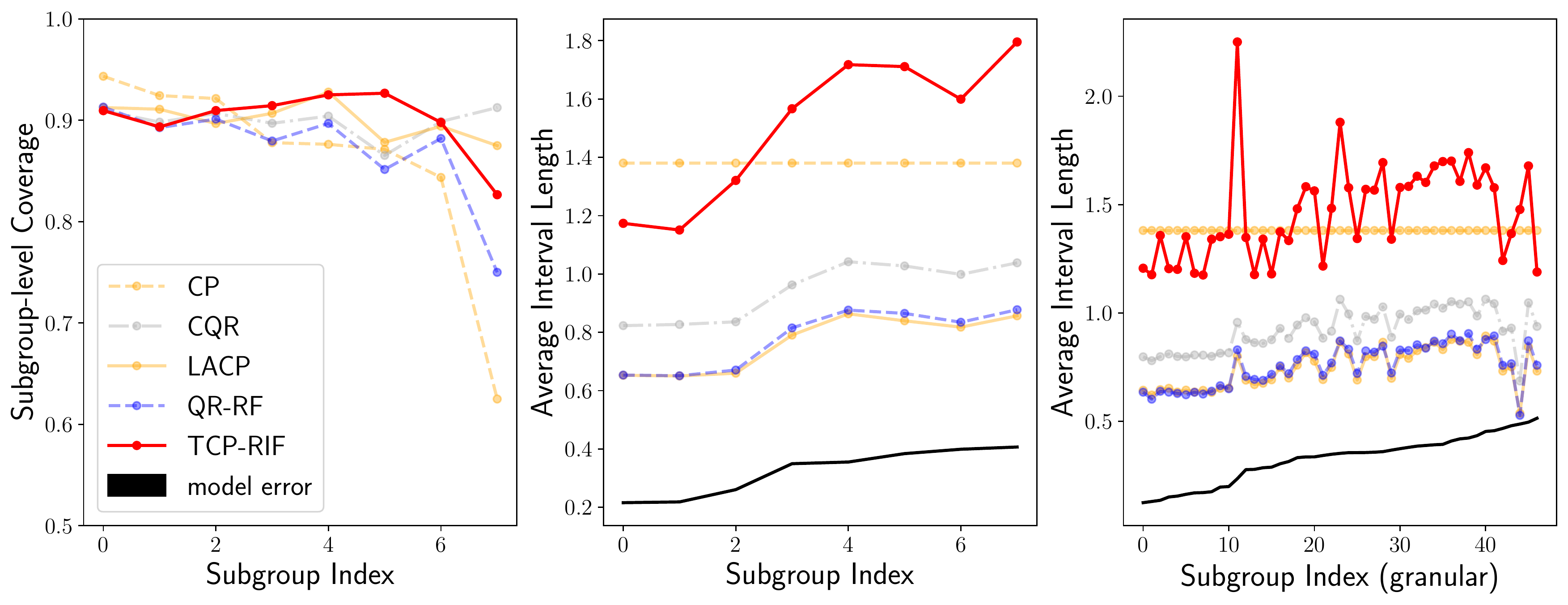}
    \includegraphics[scale=0.38]{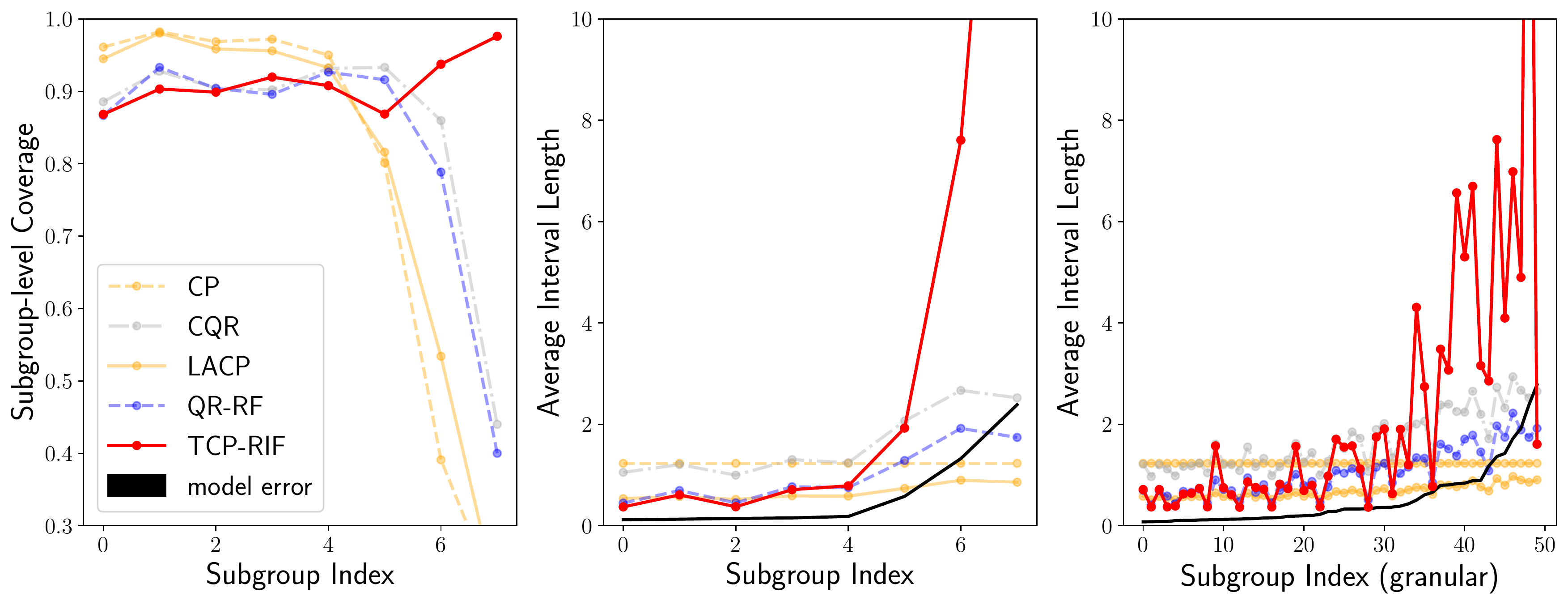}
    \caption{\footnotesize Adaptivity of predictive inference baselines on the {\bf Facebook-1} and {\bf MEPS-20} dataset.}
    \label{fig:fb}
    \rule{\linewidth}{.75pt}   
\end{figure}

\textbf{Evaluating Adaptivity} Similar to the experiment described in the main paper, we continue to evaluate the extent to which baselines are adaptive, i.e. how congruent the lengths of the intervals are with the true uncertainty of the base model, as reflected by the model error. We run this experiment on the \textbf{Facebook-1} and \textbf{MEPS-20} datasets. We see, as before, that either the baseline is not adaptive (e.g. CP or LACP) or has limited adaptiveness, but \textit{not} enough to maintain the target coverage in the subgroup (e.g. QR-RF, CQR). QR-RF and CP, in particular, have a steep dropoff in coverage for the subgroups in which the model is more uncertain (i.e. higher indexed subgroups). On the other hand, we see CUQR both maintaining coverage in all subgroups and adapting the interval lengths given the uncertainty in the subgroup.

\mycomment{
\subsection*{Appendix E: Results on the SEER Breast Cancer Dataset}
\begin{table}
\centering
{\footnotesize
\begin{tabular}{llll}
\toprule
\toprule
                 \midrule 
               & $C_{av}$     & $L_{av}$     & $C^{w.c.}_{G}$ \\ \midrule
\textbf{QR methods} &&& \\
\quad QR-RF   & 0.93\tiny{$\pm$0.00} & 10.15\tiny{$\pm$0.03} & 0.92\tiny{$\pm$0.00}\\
\quad QR-NN   & 0.80\tiny{$\pm$0.01} & 7.48\tiny{$\pm$0.61} & 0.78\tiny{$\pm$0.01}  \\
\midrule
\textbf{CP methods} &&& \\
\quad CP   &  0.90\tiny{$\pm$0.00} & 17.67\tiny{$\pm$0.05} & 0.90\tiny{$\pm$0.00}\\
\quad LACP  & 0.90\tiny{$\pm$0.00} & 8.68\tiny{$\pm$0.09} & 0.89\tiny{$\pm$0.00}\\
\quad CQR   & 0.90\tiny{$\pm$0.00} & 10.04\tiny{$\pm$0.03} & 0.90\tiny{$\pm$0.00} \\
\quad CCH   & 0.89\tiny{$\pm$0.00} & 8.20\tiny{$\pm$0.05} & 0.88\tiny{$\pm$0.02} \\ 
\quad CQ  & 0.89\tiny{$\pm$0.00} & 17.53\tiny{$\pm$0.06} & 0.90\tiny{$\pm$0.00}  \\\midrule
\quad CUQR & 0.90\tiny{$\pm$0.00} & 17.89\tiny{$\pm$0.16} & 0.90\tiny{$\pm$0.00}  \\
\bottomrule
\bottomrule
\end{tabular}}
\vspace{.05in}
\caption{\footnotesize Marginal coverage, efficiency and conditional coverage of all baselines on the SEER breast cancer dataset.}
  \vspace{-2mm} 
  \rule{\linewidth}{.75pt}   
\label{tab:sup_table_seer}
\end{table}

\textbf{Setup} 

\textbf{Coverage and Efficiency}

\textbf{Qualitative Analysis}}

\end{document}